\pdfoutput=1
\let\mypdfximage\pdfximage
\def\pdfximage{\immediate\mypdfximage}

\documentclass[12pt]{article}
\usepackage{epsfig}
\usepackage{graphicx}
\usepackage{amsmath}
\usepackage{amssymb}
\usepackage{mysubfigure}
\usepackage{array,multirow}
\usepackage{bibspacing}
\usepackage{multirow}
\usepackage{authblk}
\usepackage{appendix}
\usepackage[pagebackref=true,breaklinks=true,letterpaper=true,colorlinks=false,bookmarks=false]{hyperref}

\usepackage[letterpaper, margin=1in]{geometry}
\begin{document}

\title{Detangling People: Individuating Multiple Close People\\and Their Body Parts via Region Assembly}

\author[1]{Hao Jiang}
\author[2]{Kristen Grauman}
\affil[1]{Computer Science Department, Boston College, USA}
\affil[2]{Department of Computer Science, University of Texas at Austin, USA}
\date{}

\maketitle

\begin{abstract}
Today's person detection methods work best when people are in common upright  poses and 
appear reasonably well spaced out in the image.  However, in many real images, that's not 
what people do.  People often appear quite close to each other, e.g., with limbs linked or 
heads touching, and their poses are often not pedestrian-like.  We propose an approach 
to detangle people in multi-person images.  We formulate the task as a region assembly 
problem.  Starting from a large set of overlapping regions from body part semantic 
segmentation and generic object proposals, our optimization approach reassembles those pieces
 together into multiple person instances.  It enforces that the composed body part regions 
of each person instance obey constraints on relative sizes, mutual spatial relationships, 
foreground coverage, and exclusive label assignments when overlapping.  Since optimal region 
assembly is a challenging combinatorial problem, we present a Lagrangian relaxation method 
to accelerate the lower bound estimation, thereby enabling a fast branch and bound solution 
for the global optimum.  
As output, our method produces a pixel-level map indicating both 1) the body part 
labels (arm, leg, torso, and head), and 2) which parts belong to which individual person.
Our results on three challenging datasets show our method is robust to clutter, occlusion, 
and complex poses.  It outperforms a variety of competing methods, including existing 
detector CRF methods and region CNN approaches. In addition, we demonstrate its impact on 
a proxemics recognition task, which demands a precise representation of ``whose body part is where" 
in crowded images.

\end{abstract}

\section{Introduction}

Person detection has made tremendous progress over the last decade~\cite{piotr-survey2012}.  
Standard methods work best on pedestrians: upright people in fairly simple, predictable poses, 
and with minimal interaction and occlusion between the person instances.  Unfortunately, people 
in real images are not always so well-behaved!  Plenty of in-the-wild images contain multiple 
people close together, perhaps with their limbs intertwined, faces close, bodies partially occluded, 
and in a variety of poses.  A number of computer vision applications demand the ability to 
parse such natural images into individual people and their respective body parts---for example, 
fashion~\cite{berg-fashion}, consumer photo analysis, predicting inter-person 
interactions~\cite{uci-dataset}, or as a stepping stone towards activity recognition, 
gesture, and pose analysis.

Current methods for segmenting person instances~\cite{ff1,ff2,bo,luck,bmvc2011,poselet1,poselet2} 
take a top-down approach.  First they use a holistic person detector to localize each person, and 
then they perform pixel level segmentation.  Limited by the efficiency and performance of person 
detectors, such methods are slow when dealing with people at unknown scales and orientations.  
Furthermore, they suffer when presented with close or overlapping people, or people in unusual 
non-pedestrian-like body poses \cite{uci-dataset}. 

\begin{figure}[t]
\centering
\includegraphics[width=\linewidth]{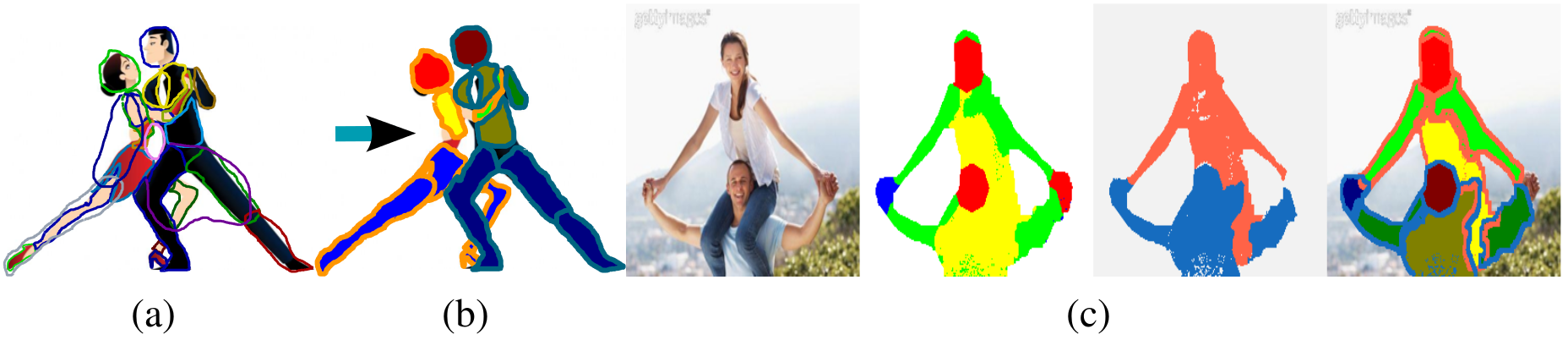}
\label{fig:teaser}
\caption{
\small
From a pool of initial regions in (a), we select a subset and assemble them
into separate human instances in (b).
Each subject's region and part boundary is marked with a different color.
Our method finds human instances and the body part regions
(arms, legs, torso
and head).
(c): Example result of our method.  From left to right: input image, semantic body part segmentation from a CNN,
person instance segmentation, final person individuation and part labeling.
(Figures in the paper are best viewed on pdf).
}
\end{figure}

We propose a new approach to detangle people and their body parts in multi-person images.  
Reversing the traditional top-down pipeline, we  pose the task as a region assembly problem 
and develop a bottom-up, purely region-based approach.   Given an input image containing an 
unknown number of people, we first compute a pool of regions using both body-part semantic 
segmentations and object proposals.  Regions in this pool are often fragmented body parts 
and often overlap.  Despite their imperfections, our method automatically selects the best 
subset and groups them into human instances.  See Fig.~\ref{fig:teaser}(a,b).  
To solve this difficult jigsaw puzzle, we formulate an optimization problem in which parts are 
assigned to people, with constraints preferring small overlap, correct sizes and spatial 
relationships between body parts, and a low-energy association of body part regions to 
their person instance.  We show that this problem can be solved efficiently 
using decomposition and a branch and bound method.

Fig.~\ref{fig:teaser}(c) shows an  example result from the proposed method.  
Note that we not only estimate pixel-level body part maps, but we also indicate 
``which part belongs to whom", even in a crowded scene with occluding people.

Experiments on three datasets show our method strongly outperforms an array of existing 
approaches, including bounding box detectors, CNN region proposals, and human pose detectors.  
Furthermore, we show the advantage of the proposed optimization scheme as compared to 
simpler inference techniques.  Finally, we demonstrate our person detangler applied to 
\emph{proxemics recognition}~\cite{uci-dataset,visual-phrase,chu}, where fine-grained estimation 
of body parts 
and body part owners is valuable to describe subtle human interactions (e.g., is he holding her 
hand or her elbow?).  

\subsection{Related work}

Most previous methods for human instance segmentation
require a person detector.
In \cite{cvpr03} a multi-part pedestrian detector is combined with an MCMC method
for walking crowd segmentation. A pedestrian detector is used in
\cite{ff1,ff2,bo} to find people instances in bounding boxes before instance segmentation.
Joint pose estimation and segmentation for single subjects have been proposed
in \cite{joint-seg-pose,joint-seg-pose2,posecut}.
Multiple people instance segmentation in TV shows has been studied in
\cite{luck,bmvc2011}
using the detector CRF scheme,
which combines a person detector and a pixel-level CRF to achieve accurate results.
In \cite{bmvc2011}, people detection bounding boxes are verified using face detections and then
grabcut is used to refine the instance segmentation.
In \cite{luck} a pictorial structure method \cite{nbest} is
used to detect candidate human instances.
Sequential assignment is used to fit the human instance masks
to image data.
From instance masks, detailed human segmentation and body part regions are
further estimated using the CRF.

Whereas existing methods largely take the strategy of first
detecting individual people and then segmenting their parts,
we propose a reversal of this conventional pipeline.
In particular, we propose to start with a pool of regions that are segments
or sub-regions of body parts on
multiple people, and then jointly assemble them into
individuated person segments.  
The advantage of not depending on a holistic person detector is
not only because these detectors have high computational complexity (especially if
we have to handle people with unknown orientations and scales),
but also
because
it is still a difficult problem for person detectors to deal with complex human poses,
inter-person interactions, and large occlusions, which our target images may include.
Compared to
previous detector-based methods,
our approach is more efficient and gives
better results.  

Deep learning approaches have been studied in the joint detection and segmentation scheme
\cite{berk1}, related to RCNN~\cite{rcnn}, though the authors target generic PASCAL object detection as opposed 
to person individuation and body part labeling.  Their method starts from object region
proposals such as \cite{derek,select,mcg}, and each region is classified as
a target, such as a human subject,
by using features on both color
images and binary image masks.
Potentially, such a method can be scale and rotation invariant
and fast. The challenge is how to propose complete whole object regions,
such as the whole mask of a person.
This is often a difficult task due to the variation of color and texture
on a single subject, the thin structure of human limbs, and arbitrary human poses.
Our proposed method also uses region proposals, but our method allows fragmented sub-regions and is able to reassemble
the broken regions back to human body parts. 

Part voting approaches have been intensively studied for human or object
instance segmentation.
In \cite{mm07}, boundary shape units vote for the centers of human subjects.
In \cite{poselet1,poselet2},
the poselets vote for the centers of people instances.
The poselets that cast the votes are then identified
to obtain the object segmentation. In \cite{reverse}
the object boundary is obtained by reversely finding the activation parts used in the voting.
Similar to the Hough Transform, such a voting approach is more suitable to targets that have relatively
fixed shape. Our proposed method finds the optimal part assembly
using articulation invariant constraints
instead of simply voting for the person center;
it therefore
can be used to segment highly articulated human subjects.

Our method is also related to human region parsing, in that we segment and label each
person's body part regions.
Human region parsing has been mostly
studied for analyzing body part regions of a single person \cite{mori,jiang,jianbo,parsing-cvpr14}. 
To handle multiple people, in \cite{bo}
a pedestrian detector is used to find the bounding box of each single
person.
Finding people with arbitrary poses using a bounding box detector is
still a hard problem, whereas our method naturally handles multiple people with
complex interactions and poses.
Part segmentation has recently been used to improve semantic segmentation of animals in \cite{ucla}, but 
the pairwise CRF method cannot individuate multiple animal instances.
In contrast, our method is able to individuate tangled people with complex poses.

Our work is also distantly related to human pose estimation, which has been intensively studied on
depth images \cite{kinect} and on color images using
pictorial structure methods \cite{ps1,ps2,ps3} and CNNs
\cite{song,chen,deeppose}.  However, unlike our approach, human pose estimation methods usually do not directly give the instance and body part region segmentation.
Our method produces multiple human segmentations without
extracting human poses.

In summary, the main contributions of this paper are:
(1)  We tackle the new problem of multiple person instance
individuation and body part segmentation from region assembly.
(2) We propose a novel linear formulation.
(3) We propose a Lagrangian relaxation method to speed up lower bound estimation,
with which
we solve the optimization using fast branch and bound.
Our experiments show that our method is fast and effective, outperforming an array of 
alternative methods, and improving the state-of-the-art on proxemics recognition.

\section{Method}

We first overview our approach (Sec.~\ref{sec:overview}), then present the big picture formulation 
of region assembly as a graph labeling problem (Sec.~\ref{sec:labeling}).  Next we describe in detail 
how we implement the components of that formulation (Sec.~\ref{sec:details}).  And, we introduce 
our efficient  optimization approach (Sec.~\ref{sec:opt}). Finally we discuss parameter setting (Sec.~\ref{sec:params}).

\subsection{Overview} \label{sec:overview}

There are different ways to generate regions and sub-regions of human body parts.
In this paper, we start with a large set of body part region proposals
using three sources: a CNN trained for semantic segmentation of body parts (details below), 
generic region proposals~\cite{derek}, and the intersection between the previous two kinds of regions.
Some of the region proposals may already give body part regions of separate human instances,
or more likely they are partial sub-regions of body parts.
Many proposal regions do not correspond to body part regions,  or may be the union of two individuals' body parts.
Our goal is to select a subset of regions from these proposals
and reassemble them to individuate
human instances and the associated body parts.
Intuitively, a good configuration should have
arm, leg, torso, and head regions in proportional sizes, and part
regions should follow correct neighborhood relations.

We show how to search for the best jigsaw puzzle assembly meeting those criteria.
In particular, we search for the optimal set of regions
 so that
they minimize an objective function.
Let $P$ be the set
of overlapped regions or sub-regions of different body parts.
Let $\mathcal{X}$ be a vector of
integers that indicate a specific region in $P$ is assigned
to a person $i$ from $1,\dots,N$ and $N$ is
the number of human instance candidates
determined by the algorithm during the optimization (details below).  The element of $\mathcal{X}$ is
zero if the corresponding region candidate does not belong to any person and a natural number
otherwise. We find the optimal $\mathcal{X}$ by jointly optimizing over all potential people instances:
\begin{equation}
\mathcal{X}^\ast = \mbox{argmin}_\mathcal{X}  
\{U(\mathcal{X}) - R(\mathcal{X}) + S(\mathcal{X})\} \label{eq:overall} 
\;\; \mbox{  s.t. }  I(\mathcal{X}) \le 0, G(\mathcal{X}) \le 0, W(\mathcal{X}) \le 0.
\end{equation}
Here $U$ is the cost of assigning part regions to
specific human instances.
$R$ is a term that encourages the selected region
candidates to cover corresponding body part regions.
$S$ is a term that enforces the assembled body regions in each detected human instance
to have correct sizes.
Apart from these terms, we also introduce constraint $I$
to limit the intersection area between the selected regions,
and $G$ to constrain the color histogram between specific region pairs.
We also use constraint $W$ to
enforce the total body part area of each instance
person to be within an upper bound. All these terms
are defined in detail below.  

\subsection{Region assembly as a graph labeling problem}\label{sec:labeling}
\begin{figure}[t]
\centering
\includegraphics[width=0.75\linewidth]{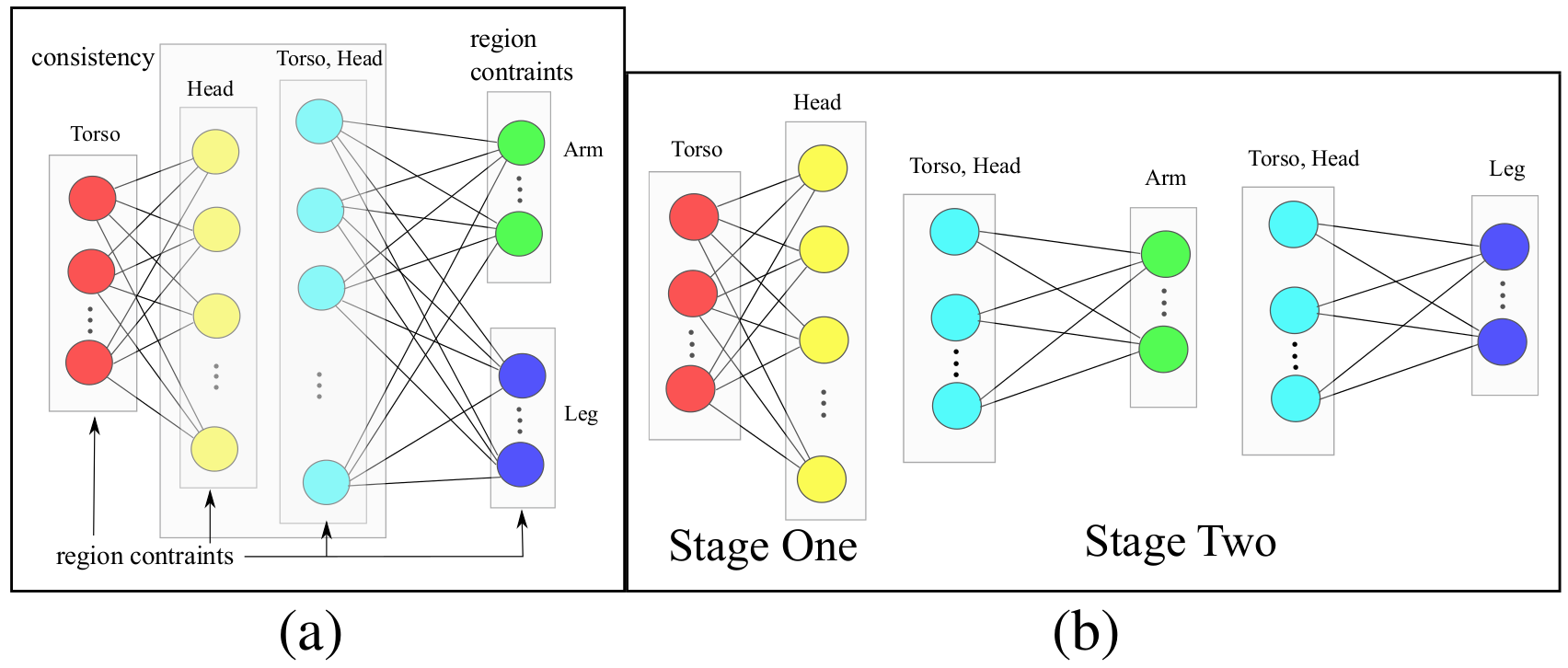}
\caption{
 \small 
(a) To optimize region assembly we find the node and edge 0-1 assignment
which minimizes the objective in Eq.~\ref{eq:overall} while satisfying different
region constraints on body part assembly. (b) We decompose the optimization into three optimizations in
two stages.  See text for details.
  }
\label{fig:ink}
\end{figure}

Fig.~\ref{fig:ink}(a) illustrates the region assembly problem
as graph labeling.
The nodes correspond to the regions or sub-regions of different body parts.
Head nodes and head-torso nodes in Fig.~\ref{fig:ink}(a)  
are also denoted as human instance nodes.
The head-torso nodes represent the head-torso region combinations.
The binary edges correspond to possible region-to-human instance assignments,
and the hyper edges constrain the region coupling and assignment consistency.
The binary edges and nodes have weights.
We essentially need to find an optimal node-edge labeling to minimize the total weight.
The optimization is combinatorial. It is hard to solve due to the large number
of edges, loopy structure and high order constraints.
Instead of directly solving the hard problem, we decompose it
into three optimizations on three simpler graphs
in two stages, as shown in Fig.~\ref{fig:ink}(b).

The three optimizations share a common integer program format:
\begin{align}
& \min_{x,y,e} \{ g^Tx + w^Ty + \phi 1^Te \}  \label{eq:cm} \\ \nonumber
& \mbox{s.t. } Ax \le 1, Bx + Ce + Dy \le f, \; e \ge 0, \; x, y \mbox{ are binary. }
\end{align}
Here, the vector $x$ includes the edge variables and the vector $y$ includes the human instance
node variables.  The dimension of $x$ equals the number of torso regions in stage one or
number of arm or leg regions in stage two times the number of candidate head regions.
The dimension of $y$ equals the number of head regions. 
$e$ is an auxiliary
variable vector.
$g, w$ are constant coefficient vectors. 
$\phi$ is a constant. $1$ is an all-one vector.
$Ax \le 1$ is the assignment constraint
and $Bx + Ce + Dy \le f$ represents the region coupling constraints.
How $g,w,f,A,B,C,D$ are determined will become clear after Sec.~\ref{sec:details}.

The optimization
finds the pairing of nodes in each augmented bipartite graph
in Fig.~\ref{fig:ink}(b). The nodes on one side
are the regions of torsos, arms or legs. The nodes on the other side
are the human instance
representations using head regions (stage one) or a head-torso region combination (stage two).
Each part region (arm, leg, torso) node can be used at
most once, and each human instance node may receive zero or multiple region matches.
When selecting multiple part nodes, we assemble corresponding body part regions using ``broken'' region pieces.
The nodes of part regions
are coupled by the size and exclusion
constraints.

\subsection{Detailed formulation}\label{sec:details}

Now we flesh out how we instantiate the general formulation presented above.  
We start from a semantic segmentation map in which each pixel is
classified as one of the four part types (arms, legs, torso and head) or the background.  
(``Background" = ``not any person"; all person pixels are ``foreground".)
The map is obtained by
first computing a stack of probability maps from a CNN (a modified AlexNet \cite{alexnet}) for each part at
different scales.
Max-pooling is then applied to compute the body part soft semantic map.
We use graph cuts with alpha-expansion to generate the final semantic segmentation map. Please see Appendix~\ref{app:cnn}
for details.

Overall, the goal is to have a large pool of part candidates with high recall, but possibly 
low precision; that way, there is a high chance that
we can correctly use them to assemble and separate multiple human human instances.  
With this in mind, regions and sub-regions of body parts (torso, arm, and legs) are generated as 
follows.
Apart from using the connected components of body part regions from the CNN-derived semantic segmentation map,
we use region proposals from \cite{derek}
to ``chop'' possibly merged part regions into smaller pieces by intersecting
the region proposals with each part region.
The regions therefore include both whole body parts and fragments of body parts.

Head regions are generated differently because the above method may not always be
able to separate close head regions.   
The head regions are circular regions whose radius
is determined by the max-response scale
at each head point; the head points are detected by finding peaks in the soft semantic head
map using non-maximum suppression. (More details in Appendix \ref{app:cnn}.)
While our framework allows multiple head candidates with different scales at the same point, 
in practice we find selecting the single most likely head candidate at each point is 
sufficient, and so we take this simpler configuration.   
The head candidate regions are further intersected with the person 
foreground in the semantic map. 
The head detections automatically tell our system the candidate people number in the image. 

Now we are ready to define the energy terms $U$, $R$, $S$ and constraints $I$, $G$, $W$ 
in Eq.~\ref{eq:overall} 
and derive Eq.~\ref{eq:cm}.
We introduce a binary variable $x_{i,j}$, the binarized version
of $\mathcal{X}$ in Eq.~\ref{eq:overall},  
to label edges in Fig.~\ref{fig:ink}(b):
$x_{i,j} = 1$ if region $i$
is selected to
be part of person $j$, otherwise $x_{i,j} = 0$.
We have the following constraint on $x$:
$\sum_{j} x_{i,j} \le 1$, which means
each region can only be assigned to
at most one human instance.
Each person instance may connect to multiple
regions to handle the region splitting case.
We also introduce variable $y_j$ to indicate whether person/head candidate $j$ is
selected.  $y$ is the human instance node variable. We enforce
$y_j \ge x_{i,j}, \forall i$.  In stage two, the head-torso regions come 
from the solution of stage one, and $y$ is all one.

\subsubsection{Region assignment costs $U$:}

There is a cost $c_{i,j}$ to associate region $i$
to person instance candidate $j$, and a cost $p_j$ to select instance candidate $j$.  
The total assignment cost
is
$U(\mathcal{X}) = \sum_{i,j} c_{i,j} x_{i,j} + \xi \sum_j p_j y_{j}$.
We optimize $y$ only in stage one. In stage two,
$y$ is fixed to be all ones and can be
removed from the optimization.
$p_j$ equals
one minus the head region's peak probability on the head map, so as to emphasize costs incurred on more confident heads. $\xi$
is a constant weight balancing the region association cost against instance selection cost.

In short, the cost $c_{i,j}$ aims to associate a person instance with regions that
``look like'' part of the corresponding body parts, mostly fall in the range radius, and are close to
the human instance. We also encode the preference to use a small set of large regions to build the body parts
of each human instance, to avoid trivial assemblies (e.g.,~\cite{irani-smallsegs}).

Specifically, the assignment cost $c$ is defined as:
\begin{equation}
c_{i,j}=\alpha q_{i,j}+\beta r_{i,j} + \gamma d_{i,j} + \theta,
\end{equation}
where $q_{i,j}$ equals one minus the mean probability of region $i$ belonging to human instance $j$ as
part of a specific body part
(probability obtained from the soft CNN semantic maps at $j$'s scale).
The term $r_{i,j}$ is the proportion of region $i$ exceeding the range radius
of instance $j$. To maintain scale-invariance, the range radius for each body part is estimated using the
instance/head $j$'s scale and a reference person height (150 pixels).  
The term $d_{i,j}$ is the 
shortest point-to-point distance between the head region (stage one) or head/torso region (stage two)
$j$ and part region $i$, normalized based on the reference height.
The constant $\theta$ penalizes selecting many small regions to construct a body part
in each human instance.
The coefficients $\alpha$, $\beta$, $\gamma$ and $\theta$ control the weights
among different terms. The setting of these coefficients are discussed later.

\subsubsection{Size term $S$ and constraint $W$:}

When composing a human instance's body part, the total area of
the selected regions is limited by the body part's size: $\sum_{i} a_{i} x_{i,j} \le s^2_j b$,
where $a_{i}$ is the area of the region $i$, $s_j$ is the scale of the human instance $j$
and $b$ is largest possible area of a body part
for the reference person (150-pixel tall). In stage one $b$ is the max area of the torso, 
and in stage two $b$
limits the area of arms or legs. 
Apart from the hard constraint, a soft one encourages the total area
of a region assembly
to approach a target size of the corresponding body part.
We minimize $|(\sum_{i} a_{i}/s^2_j x_{i,j}) - l|$, which
can be converted to a linear form:
$\min e_j, \mbox{ s.t. } -e_j \le (\sum_{i} a_{i}/s^2_j x_{i,j}) - l \le e_j, e_j \ge 0$.
Here $l$ is the 
the average body part size of 
the reference person from different view points.
It corresponds to the torso in stage one and arms or legs in stage two.

\subsubsection{Exclusion and color consistency constraints $I$ and $G$:}

We also prefer to select regions that are mostly non-overlapping to form each body part region.  Thus, we introduce an exclusion 
constraint $I$ to discourage overlap.
Let $z_i = \sum_j x_{i,j}$ indicate whether region $i$ is associated
to a human instance. To construct constraint $I$, we let
$z_m + z_n \le 1, \; \mbox{if } q_{m,n} > \tau$,
where $q_{m,n}$ is the area intersection to union ratio between region $m$ and $n$, $\tau$ is a constant.

Apart from intersection exclusion, we
also prefer that the color histograms should match if two regions are
selected to form the same body part.
We thus enforce the constraint $G$ that $x_{u,j} + x_{v,j} \le 1, \; \mbox{if } h_{u,v} > \varepsilon$, where $h_{u,v}$
is the $L_1$-distance between the normalized color histogram of region $u$ and $v$, $\varepsilon$ 
is a constant threshold.

\subsubsection{Max covering term $R$:}

If we simply minimize the above terms, $x, y$ will be all zero since all
the coefficients in the objective are non-negative. We introduce an extra covering term
to encourage the chosen regions to cover the corresponding body part regions in the semantic segmentation map.
We maximize the total region size $\sum_{i} r_i z_i$, where $r_i=a_i/m_{t_i}$
and $t_i$ is the part type of candidate $i$, $m_{t_i}$ is the
total area of part $t_i$ in the semantic map.
$R$ is proportional to the total region size.
This encourages region covering
because we enforce the regions to be mostly disjoint.

Combining the above terms, we have our final optimization objective:
\begin{align}
& \min \left\{\sum_{i,j} c_{i,j} x_{i,j} + \xi \sum_{j} p_j y_j + \phi \sum_j e_j - \pi \sum_{i} r_i z_i \right\} \\ \nonumber
& \mbox{s.t. } \sum_{j} x_{i,j} \le 1,~~~ z_i = \sum_j x_{i,j},
       y_j \ge x_{i,j}, \forall i,j \\ \nonumber
     & z_m + z_n \le 1, \; \mbox{if } q_{m,n} > \tau,
     ~~~x_{u,j} + x_{v,j} \le 1, \; \mbox{if } h_{u,v} > \varepsilon \\ \nonumber
     & \sum_{i} a_{i} x_{i,j} \le s^2_j b, \; -e_j \le (\sum_{i} a_{i}/s^2_j x_{i,j}) - l \le e_j, e_j \ge 0,
\end{align}
where $\phi$ and $\pi$ are coefficients that serve to control the weights of size term and cover term. 
It is easy to verify that the formulation
indeed has the format in Eq.~\ref{eq:cm}, if we vectorize
variables $x, y, e$ and substitute $z$ by $x$ terms.
The optimization is a hard combinatorial problem due to the loopy structure and
high order constraints.
We next propose
an efficient relaxation and branch and bound method to solve the problem.

\subsection{The lower bound}\label{sec:opt}

A direct linear relaxation, in which
we replace the binary constraints on $x$ with a 0 to 1 soft constraint,
gives a good lower bound.
 The disadvantage of
this approach is that we need a linear programming solver, and as the
number of region candidates increases the complexity of solving a large
linear program is high. With 1000 candidates and 2 human instances,
the simplex method takes around 4 seconds to complete, while using the following
speedup the time can be reduced to 0.1 seconds using the same CPU. 

We obtain the lower bound using the Lagrangian dual.
The size constraints and the exclusion constraints complicate the problem.
We move them into the objective function.
To simplify notation we use the compact
format of Eq.~\ref{eq:cm}:
\begin{align}
& \max_{\nu}\min_{x,y,e}  \{g^Tx + w^Ty + \phi 1^Te + \nu^T(Bx + Ce + Dy - f)\} \nonumber\\
& \mbox{s.t. } Ax \le 1, 0 \le e \le M, ~~x,y \mbox{ are binary}, \nu \ge 0,
\end{align}
where $\nu$ is the Lagrangian multiplier vector.
We introduce an upper bound $M$ for $e$ to avoid unbounded solutions.
Since the extra term in the objective is
non-positive for all the feasible solutions of the original problem,
the Lagrangian dual gives a lower bound.

The internal part of the dual is easy to solve because it can be
decomposed into three simple problems (no P2 in stage two):
\begin{align}
& \mbox{[P1]:} \min_{x} (g^T +\nu^TB)x, \mbox{ s.t. } \;  Ax \le 1, \; x \mbox{ is binary}. \\
& \mbox{[P2]:} \min_{y} (w^T +\nu^TD)y, \mbox{ s.t. } \;  y \mbox{ is binary}. \\
& \mbox{[P3]:} \min_e (\phi 1^T + \nu^TC)e, \mbox{ s.t. } 0 \le e \le M.
\end{align}
P1 can be solved by sequential assignment: in an assignment graph such as Fig.~\ref{fig:ink}(b), for each body part region node,
we check all the links to the human instance node
and find the most negative link and let the corresponding $x$ variable
to be 1. If there is no negative
link, no matching is made and the corresponding $x$ is 0.
In P2 and P3, $y$ is set to 0 or 1 and $e$ is set to 0 or $M$ according to the positiveness
of their coefficient.

Each set of Lagrangian multipliers corresponds to a lower bound of the original
problem. We are interested in the largest lower bound. The bound with respect to
the multipliers is a concave function and can be solved using the subgradient method.
The iteration alternates between solving for $x,y,e$ and updating $\nu$
by $\nu \leftarrow \max(0, \nu + \delta (Bx + Ce + Dy - f))$.
Here $\delta$ is a small constant $10^{-6}$.
The initial values of these coefficients in $\nu$ are set to zero.

For this problem, the Lagrangian relaxation bound
is the same as that of the linear program relaxation. This is due to total unimodularity
of the internal problem of the Lagrangian dual \cite{integer}.

\begin{figure}[t]
\centering
\includegraphics[width=\linewidth]{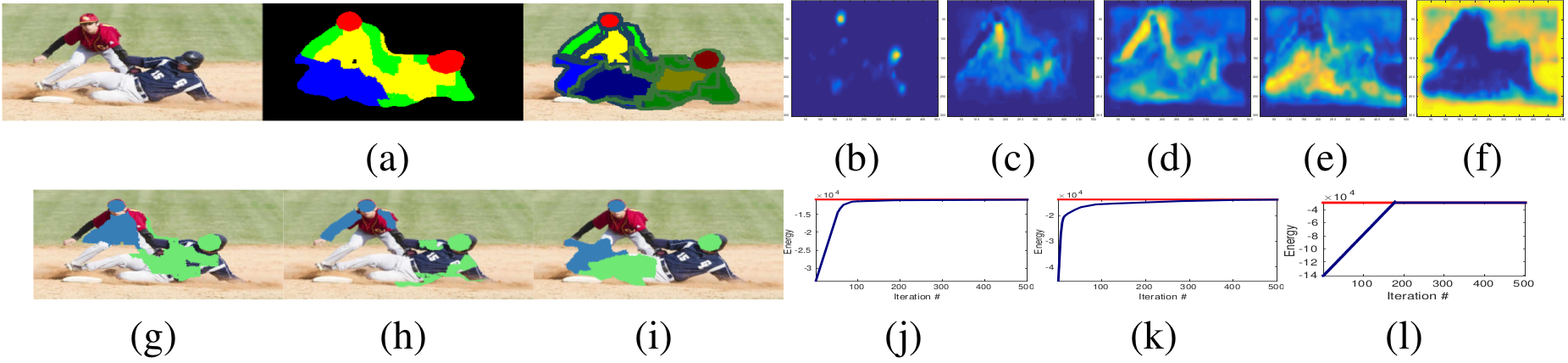}
\caption{ 
\small 
(a): From left to right: input image, semantic segmentation from CNN, and
region assembly result of the proposed branch and bound method
(shading and boundary color show the instance segmentation).  In (a), note how the CNN output does \emph{not} 
individuate parts into person instances (center), whereas our output does (right).
(b-f): Soft semantic segmentation (probability map)
from CNN for head, torso, arm, leg, and background. (g-i): Part selection
using Lagrangian dual for the torso, arms, and legs. For clarity, torso is not shown
in the stage two optimization.
Color indicates instance group. (j-l) show the energy of the Lagrangian dual
approaches the solution (red line) of the linear program relaxation.
   }
\label{fig:example}
\end{figure}

\textbf{Example.}
Fig.~\ref{fig:example} shows an example of using the Lagrangian relaxation to obtain the lower bound.
The max-pooling result of the initial CNN semantic body part segmentation is
shown in Fig.~\ref{fig:example}(b-f).
We then use graph cuts with alpha-expansion to generate the hard
semantic segmentation map
(middle of Fig.~\ref{fig:example}(a)).  Note how parts are not individuated per person in this map.
Arm, leg and torso region proposals are generated from the intersection
between semantic part maps
and object proposals.
The Lagrangian relaxation is applied to three
optimizations in two stages.
As shown in Fig.~\ref{fig:example}(j-l), the result converges quickly to the linear program
relaxation result (the red line) in a few hundred iterations.
The relaxation assignment is illustrated in Fig.~\ref{fig:example}(g-i).
We see it is indeed very similar to the globally optimal solution in
Fig.~\ref{fig:example}(a).

The complexity of finding the lower bound using the Lagrangian relaxation is $O(n)$, where $n$
is the number of region proposals times the number of human instance candidates,
and we use a fixed number
of iterations in the subgradient method.  In contrast, the average complexity of a
linear relaxation \cite{linear} using the simplex method
is $O(n\log(n))$.  The above dual approach can be extended to estimate the lower bound
at each node of the search tree. With the lower bounds, we use the branch and bound method
to find the global optimum quickly.
For most problems in the experiments, where $n$ averages around $500$,
the branch and bound
procedure terminates in a
few seconds.

\subsection{Parameter setting}\label{sec:params}
We set the thresholds $\tau=0.2$ and $\varepsilon=0.5$.
The weights of the unary cost terms are set to be
$\alpha=200, \beta=100, \gamma=100, \theta=40$ and the weights for the
energy terms as $\xi=500, \phi = 1, \pi=2\times10^5$.
We fixed all parameters for all experiments after manually inspecting a few examples.  
With more labeled data, we can optimize these
parameters for even better performance.
(Please see Appendix~\ref{app:para} for details of how we can automatically set the parameters by
maximizing the margin on positive/negative examples via a linear program.)

\section{Experimental results}
\begin{figure*}[!h]
\centering
\includegraphics[width=\linewidth, height=0.4\linewidth]{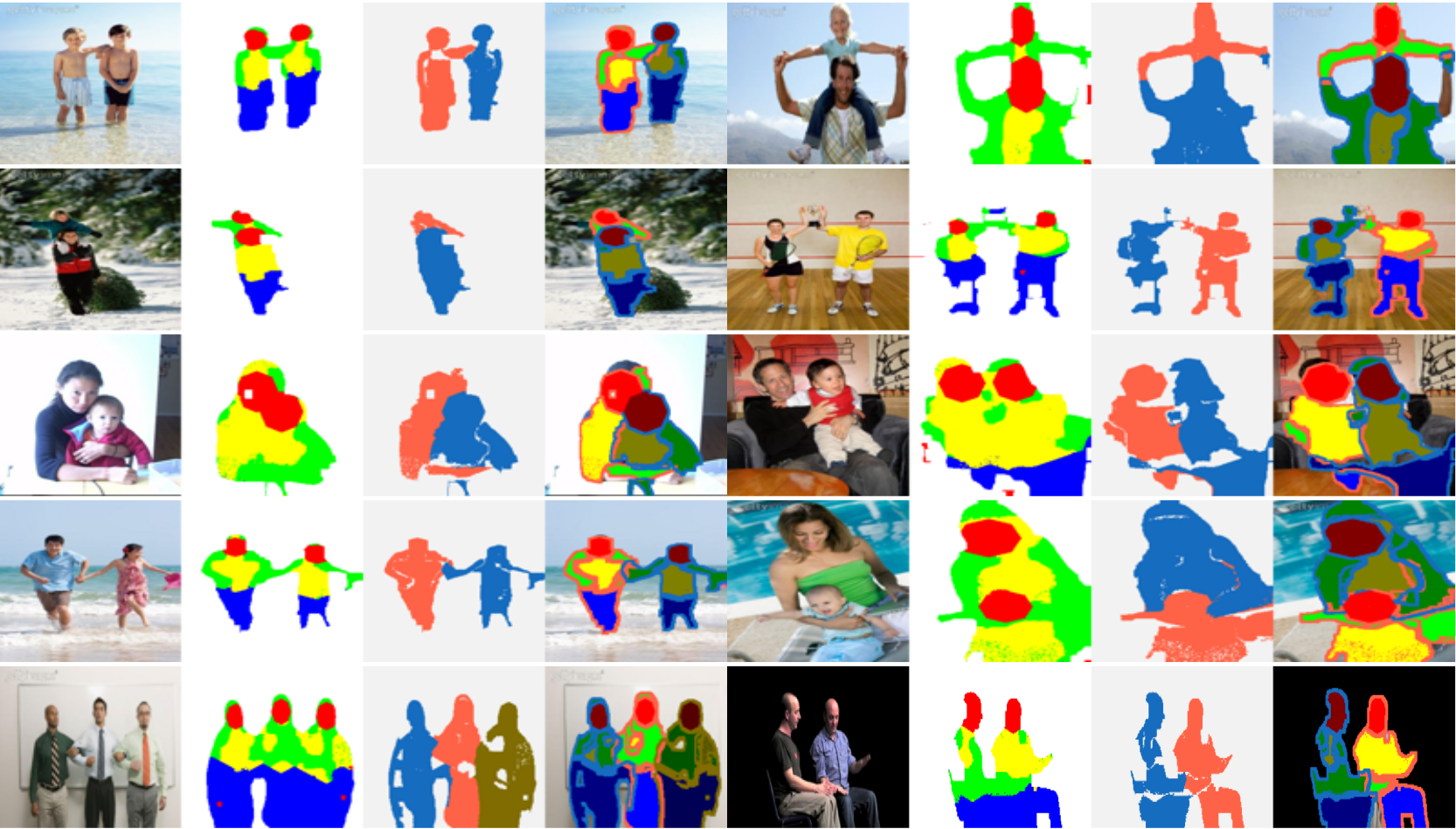}%
\linebreak
\includegraphics[width=\linewidth, height=0.4\linewidth]{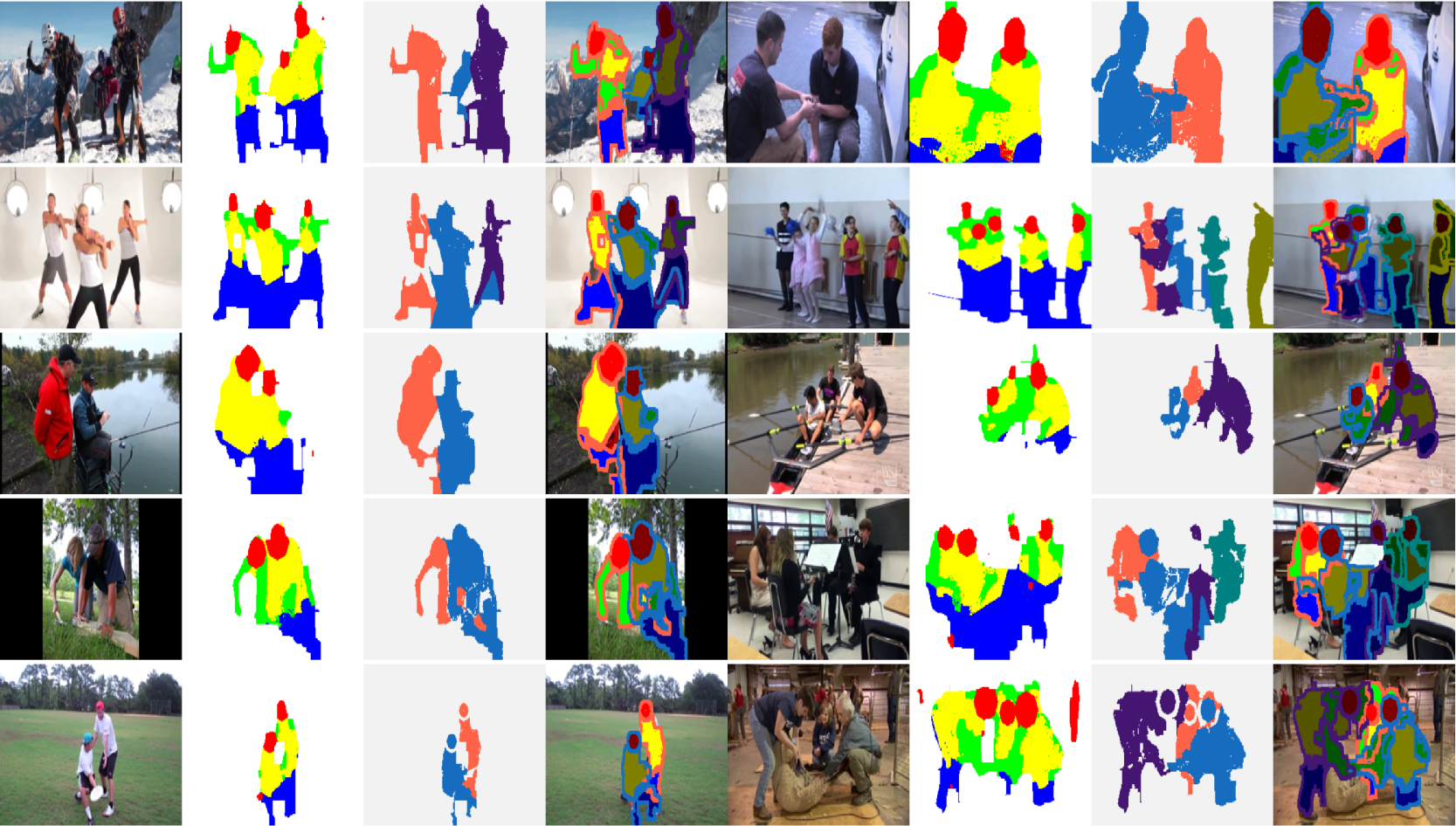}%
\linebreak
\includegraphics[width=\linewidth, height=0.16\linewidth]{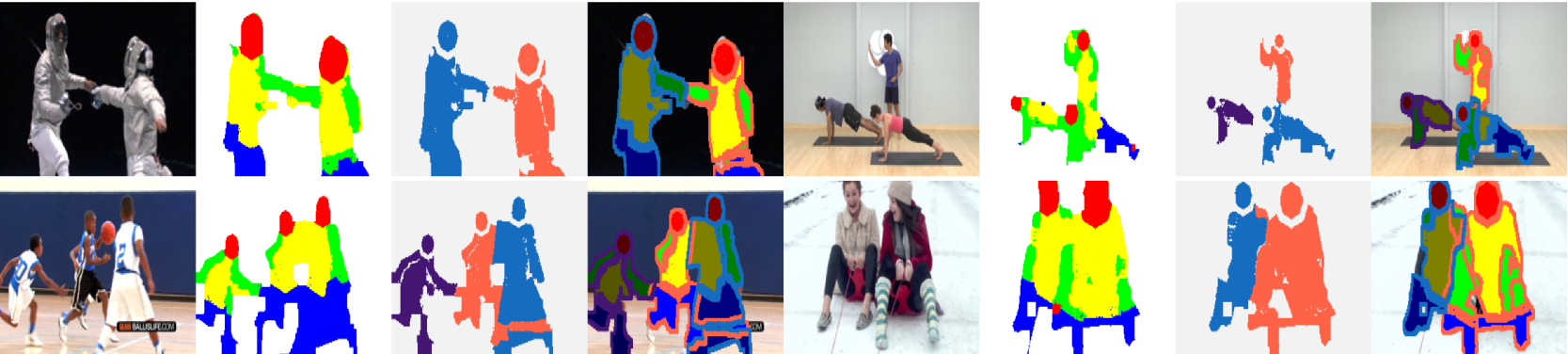}%
\caption{
\small  
Sample results on the UCI and MPII dataset.
Each result contains four columns: (1) the input images, (2) our
input semantic segmentation body part map, (3) final instance segmentation, and (4) final
body part segmentation using the proposed method.
We use both shading and different boundary colors to show the segmentation.
The same body parts have the same chromaticity (arm: green, leg: blue, torso: yellow, head: red) but have different brightness
if they belong to a different person. 
}
\label{fig:samples}
\end{figure*}

\textbf{Overview:}
In the following, we compare our approach to 1) simpler inference methods, to show the value added 
over the initial CNN body part maps; 2) bounding box detector methods; 3) CNN methods using 
region proposals; 4) human pose detection based methods.  Having established our method's accuracy, we then 
demonstrate its applicability for a downstream task: proxemics recognition.

\noindent \textbf{Datasets and evaluation metrics:}
We evaluate the proposed method on 3 datasets:
UCI~\cite{uci-dataset}, which  contains 589 images, 100 images from the MPII dataset~\cite{mp2} that contain multiple tangled people, 
and Buffy~\cite{bmvc2011}.  
The images
include complex human poses,
interactions, and occlusions among subjects.
The person scales and orientations are unknown.  We manually label the human instances and four part regions in UCI and MPII dataset 
for ground truth evaluation only (not to train the CNN).

\begin{figure}[tb]
\centering
\includegraphics[width=\linewidth]{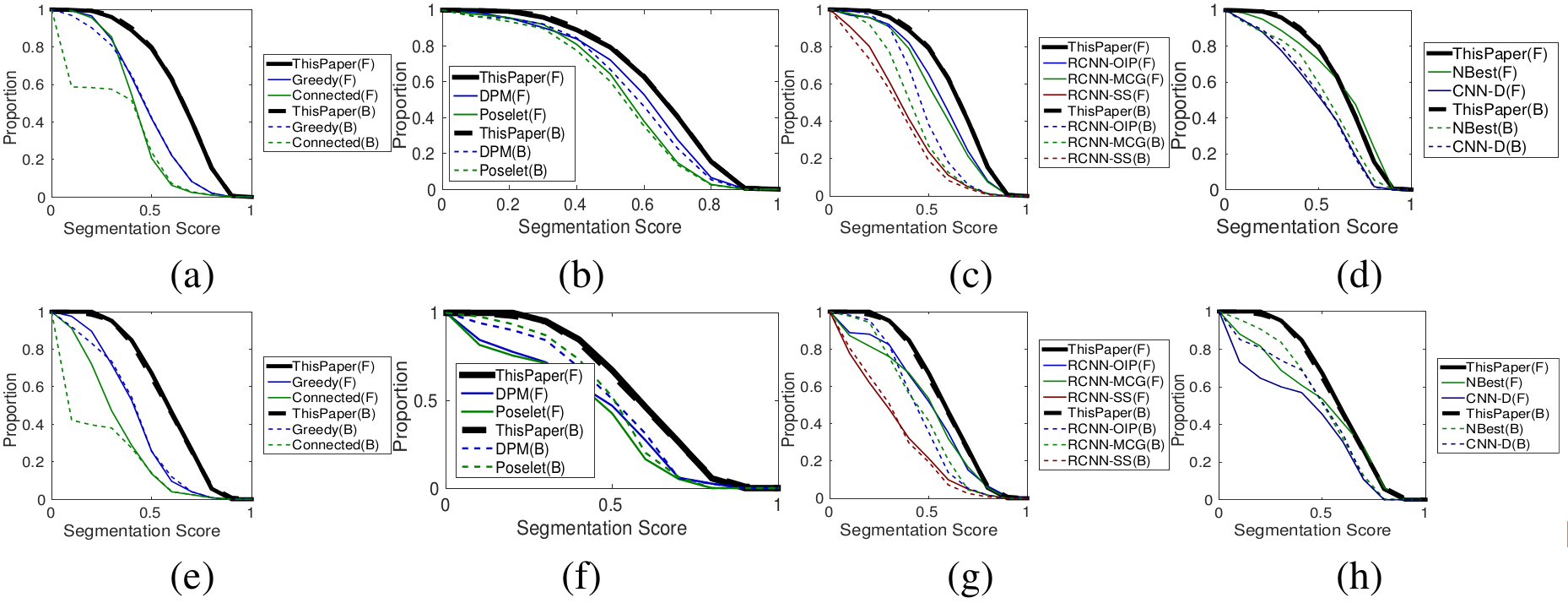}
\caption{ 
\small  
The person instance IoU ratio curves for UCI (a-d) and MPII (e-h) datasets.  We compare to 
(a, e): connected component and greedy method,
(b, f): people detectors combined with grabcut,
(c, g): RCNN with region proposal methods, and
(d, h): methods using pose detectors \cite{nbest} \cite{chen}.  F and B denote forward and backward scores.
}
\label{fig:curve}
\end{figure}

We use the area intersection to union (IoU) ratio against the ground truth labeling to quantify the performance.
We report the IoU for the human instances and mean IoU over all body part labels within each instance.
To compute forward (F) scores, we match
each ground truth segment
to the best segmentation result.  For the backward (B) scores, the matching is the other way around.
The forward score is affected by missing detections and the backward score by the false alarms.
The plots in Fig.~\ref{fig:curve} sweep through overlap thresholds and plot the proportion of results that
have instance IoU ratios higher than each threshold. The higher the curve, the better the performance.
Table~\ref{tab1} shows average IoU scores.  We explain the methods compared below.

Fig.~\ref{fig:samples} shows sample results of our method on UCI and MPII.
\begin{figure}[h!]
\centering
\subfigure[]{\includegraphics[width=0.7\linewidth]{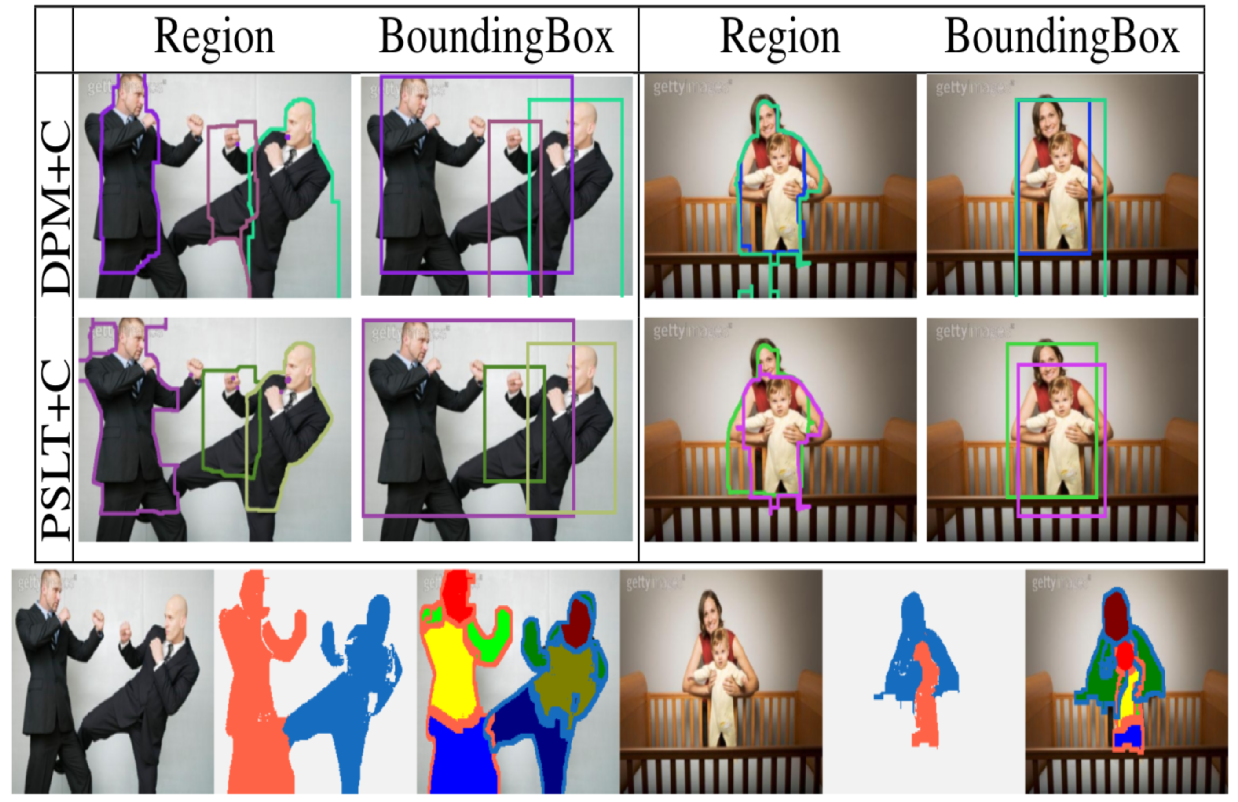}}%
\linebreak
\subfigure[]{\includegraphics[width=0.75\linewidth]{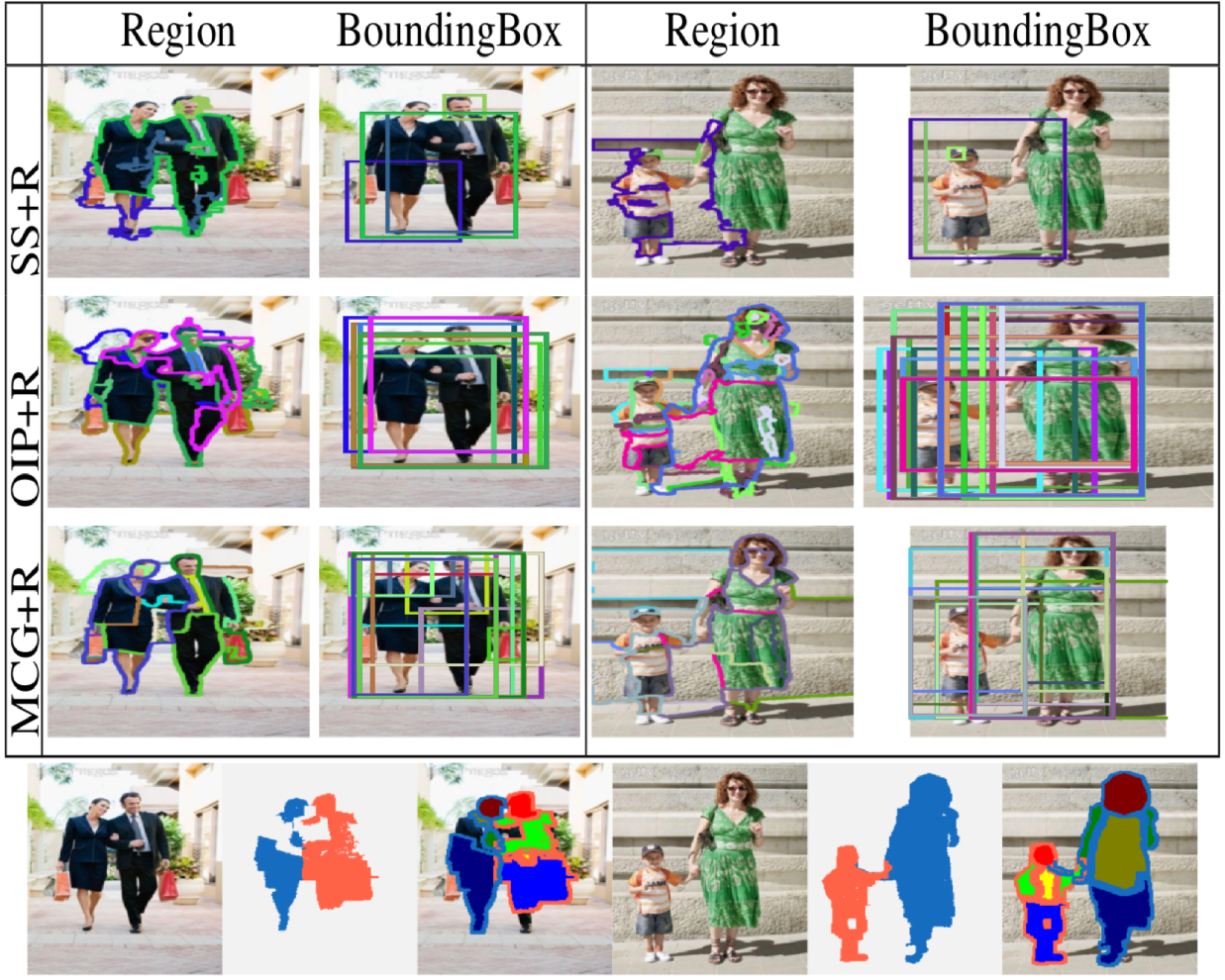}}%
\caption{
\small  
Comparison with person segmentation using DPM \cite{dpm} and Poselet (PSLT) \cite{poselet1} detectors combined
with GrabCut (C) \cite{grabcut} in (a), and object proposal methods (selective search (SS) \cite{select},
object independent proposal (OIP) \cite{derek},
and MCG \cite{mcg}) combined with an RCNN (R) person detector~\cite{rcnn} in (b).
Our results are shown in the last row of (a) and (b).}
\label{fig:pff_poselet}
\end{figure}

\begin{figure}[tb]
\centering
\includegraphics[width=0.8\linewidth]{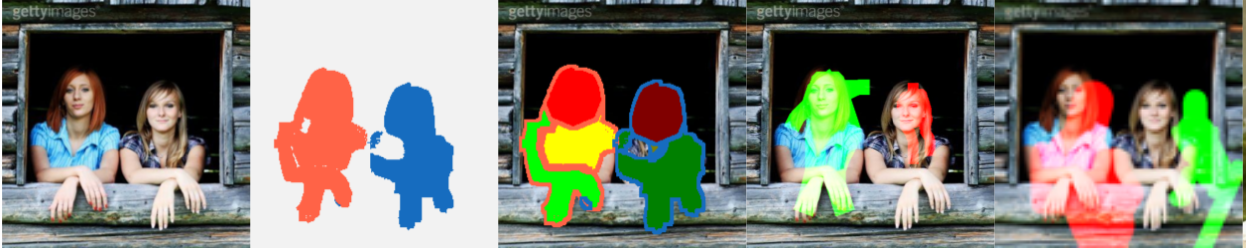}%
\linebreak
\includegraphics[width=0.8\linewidth]{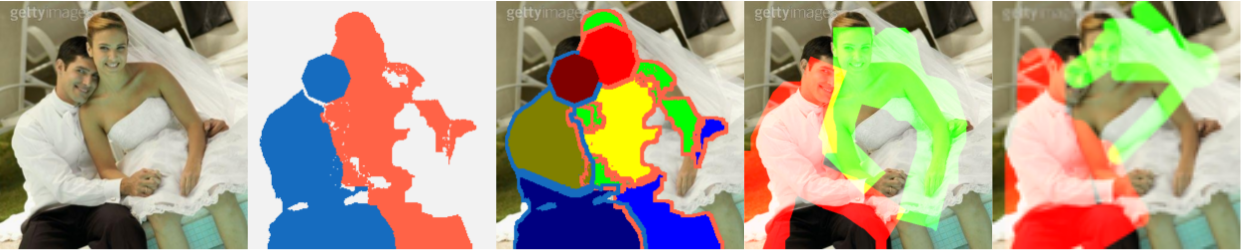}%
\caption{ 
\small  
Comparison with methods that use human pose detectors~\cite{nbest,chen}.
Our method's results are in column 2-3.
Column 4 shows results of \cite{nbest} and column 5 shows results of \cite{chen}.
Here we show the pose masks before CRF refinement.
}
\label{fig:nbest}
\end{figure}

\noindent \textbf{Are our initial CNN body part maps enough?  Would a simpler inference method on top of the CNN maps be sufficient?}
First, we stress that the CNN body part maps are not enough by definition, as they do not 
individuate which body part blobs go to which person.  The person 
and part segmentations merge when people are close.  For example, if their arms touch, that yields one 
connected component in the CNN output; see Fig~\ref{fig:samples}, second column in each set.

Our CNN semantic segmentation itself is reasonable.
On UCI and MPII, the average foreground pixel accuracy
and part pixel accuracy are
$73.13\%$ and $42.41\%$ respectively.  
However, this does not easily  transfer to a good human instance segmentation.
To confirm this quantitatively, we test 1) a baseline that returns connected components 
in the CNN map for the body part labels (\textbf{Connected}), and 2) a baseline that greedily 
finds the grouping of each person sequentially (\textbf{Greedy}). For the latter, after the 
lowest cost group is found, the regions in that group are removed and we proceed to the 
next one until all the head regions are exhausted.  
Note that naive exhaustive search is extremely slow due to the
huge search space. 

Fig.~\ref{fig:curve}~(a, e) and Table~\ref{tab1} show the results.   
Our full method's strong results relative to both these baselines reveals the role of our 
region assembly optimization.  Our efficient global optimization is necessary.

\noindent \textbf{Comparison with bounding box detector methods:} One widely used method (e.g., \cite{ff1,ff2,bo,poselet1}) 
 to extract human
instances is to first detect
people in a set of bounding boxes, and then obtain pixel-level segmentation.
To test such a baseline, we use a deformable part model (\textbf{DPM}) person detector~\cite{dpm}
and poselet (\textbf{Poselet}) person detector~\cite{poselet1}, and refine the segmentation with GrabCut~\cite{grabcut}. 
We adjust the threshold 
of the person detectors
to the lower side so that they can detect more people instances. We also
adjust the parameters of GrabCut to achieve the best performance.

As shown in Fig.~\ref{fig:pff_poselet} (a), when the people have complex
poses, interactions, and occlusions, the bounding boxes from person detectors
are not accurate.
It is a non-trivial task for a pixel-level segmentation method to correct such errors without
manual interaction.  Indeed, our method gives consistently superior results to the detector based approach (see Fig.~\ref{fig:curve}(b, f) and in
Table~\ref{tab1}).

\noindent \textbf{Comparison with CNN object detectors using region proposals:}
Another method for human instance segmentation is first generating many
region proposals and then using a classifier to extract
true human instances, e.g. \cite{berk1}.
RCNN \cite{rcnn} can also be modified to achieve such a function.
To compare this idea to our method, we test three kinds of region generation methods:
selective search~\cite{select}, MCG~\cite{mcg} and object-independent proposals~\cite{derek}.
Each rectangle image patch
that encloses a region proposal is then sent to RCNN to
determine the probability of the image patch containing a human instance.
For fair comparison, apart from the original dataset for training,
we also include the LSP images \cite{lsp} in the refinement, which improves
the baseline's human classification result. The RCNN detection threshold is set to 0.1. 

Fig.~\ref{fig:pff_poselet} (b) shows sample results.  
In images with tangled people instances, region proposals often have a hard time to
obtain full human segmentation, because the human structures are not directly
used in these region proposal methods.  Fig.~\ref{fig:curve}(c, g)
and Table~\ref{tab1} show the quantitative comparison.  Our method gives better results.
\setlength{\tabcolsep}{0pt}

\begin{table}[tb]
\centering
\scriptsize
\setlength{\tabcolsep}{3pt}

\begin{tabular}{c c}
Instance mean IoU: &
\begin{tabular}{ |c|c|c|c|c|c|c|c|c|c|c|c| }
\hline
 & 
    & Ours          & Connected & Greedy  & DPM    & Poselet  & R-I & R-II & R-III & NBest & CNN-D\\
\hline
\multirow{2}{*}{UCI}
& F &   \textbf{63.02}  & 41.62     & 46.88   & 57.64  & 53.50    & 56.04  & 54.01   & 36.32 & 61.81 & 48.58 \\
& B &   \textbf{63.45}  & 29.16     & 45.91   & 55.59  & 51.72    & 47.10    & 41.47  &  33.47 & 57.48 & 48.96\\
\hline
\multirow{2}{*}{MPII} & F &   \textbf{57.48}  & 30.88     & 40.15 & 42.21  & 40.00    & 56.04  & 54.01   & 36.32 & 47.74 & 38.24\\
& B &   \textbf{57.15}  & 18.85     & 39.88 & 47.91  & 48.43    & 47.10    & 41.47  &  33.47 & 48.66 & 45.48\\
\hline
\end{tabular} \\ \\
Part mean IoU: & 
\begin{tabular}{ |c|c|c|c|c|c|c|c|c|c|c|c| }
\hline
      & Ours(F) & Ours(B)    & C(F) & C(B) & G(F) & G(B) & NB(F)    & NB(B) & CD(F) & CD(B)\\
\hline
UCI &   \textbf{38.39}  & \textbf{38.56} & 24.75 & 18.43 & 27.29 & 32.30 & 37.98  & 31.08 & 26.49 & 26.75 \\
\hline
MPII &   \textbf{35.48} & \textbf{35.47} & 20.26 & 12.54 & 24.25 & 29.80 & 28.71  & 29.16  & 22.27 & 28.91 \\
\hline
\end{tabular}
\end{tabular}
\caption{ 
\small
Average person instance and part IoU ratio comparison for the UCI and MPII dataset. Connected: Connected component method.
R-I: RCNN+OIP,
R-II: RCNN+MCG, R-III: RCNN+SelectiveSearch, CNN-D: CNN pose detector \cite{chen}. F: forward score. B: backward score. In part IoU table: Connected component is denoted as C, Greedy method as G, Nbest as NB and CNN-D as CD. The numbers are percentages.}
\label{tab1}
\end{table}

\noindent \textbf{Comparison with methods using human pose detectors:}
Next, we compare our approach to two stick figure pose detectors (postprocessed to provide segmentations).  The first uses a flexible stick figure person detector~\cite{luck}; the second is based on CNNs for part detection~\cite{chen}.  

Since \cite{luck} does not report the human instance segmentation score
and code is not publicly available, we
compare with the upper bound performance
of the \textbf{N-best} poses \cite{nbest} that \cite{luck} uses for human segmentation.
We follow \cite{luck} to prune the N-best poses to remove very close estimates
while maintaining the variety; a few thousand candidate poses are extracted.
These poses are then refined to person masks following \cite{luck}.
Instead of selecting the best candidates using the energy as in \cite{luck}, we directly
find candidates that maximize the IoU ratio score using ground
truth. We also specify the order of the matching when computing the forward score so that
occlusion can be counted away. The score is thus an upper bound of the baseline.

The CNN pose detector~\cite{chen} baseline (\textbf{CNN-D}) is designed to detect a single person stick figure.  To make it generalize to our multi-person images,
we use DPM~\cite{dpm} to detect candidate bounding boxes and then apply the CNN pose detector~\cite{chen} to find poses
in each bounding box. We refine the stick figure detection to obtain instance segmentation following~\cite{luck}.

As seen in Table~\ref{tab1}, our method outperforms both pose-based methods (NBest and CNN-D) on UCI and MPII overall.
  Fig.~\ref{fig:curve}(d)
shows that the upper bound for NBest is
slightly more accurate in the high score range, because UCI contains
people mostly in upright standing poses.

Fig.~\ref{fig:nbest} shows samples of raw masks whose
refinement best fits the ground truth regions.
Our method is more robust when handling occlusion and complex people interactions
than traditional stick figure pose detectors.
Apart from the instance segmentation scores,
our method also gives better part segmentation scores than the
pose detector methods (see Table~\ref{tab1}).

\noindent \textbf{Comparison to state-of-the-art in person individuation:}
We compare to~\cite{bmvc2011}, a method specifically aimed at human individuation that 
represents the state of the art, with our method
on all the images from the Buffy dataset episode 4, 5, 6.
Our average forward and backward
scores are 68.22\% and 69.66\%, which are higher than the average score of 62.4\% reported in \cite{bmvc2011}.  
Note that \cite{bmvc2011} is trained on the Buffy dataset but ours is not.

\begin{figure}[t]
\centering
\includegraphics[width=\linewidth]{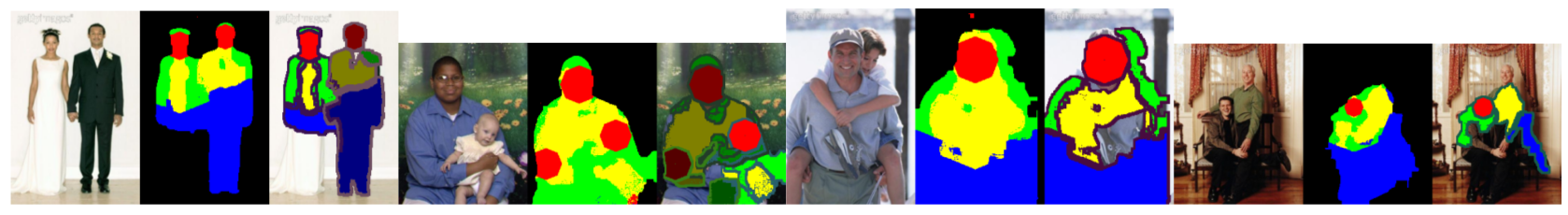}
\caption{ 
\small
Sample failure cases. }
\label{fig:failx}
\end{figure}

The detector CRF approach~\cite{luck,bmvc2011} also has higher complexity than our method, especially
when we do not know the people's orientation.
Finding a large set of pose candidates in 10 orientations alone takes
3 minutes with a 3GHz machine. Our method takes around 1 minute per image with
30 seconds on region proposal generation and less than 10 seconds on region assembly.  
Our method is rotation invariant. Our method may fail if gross errors happen on semantic maps; 
Fig.~\ref{fig:failx} shows a few failure cases.

\begin{figure}[t]
\centering
\setlength{\tabcolsep}{5pt}
\includegraphics[width=\linewidth]{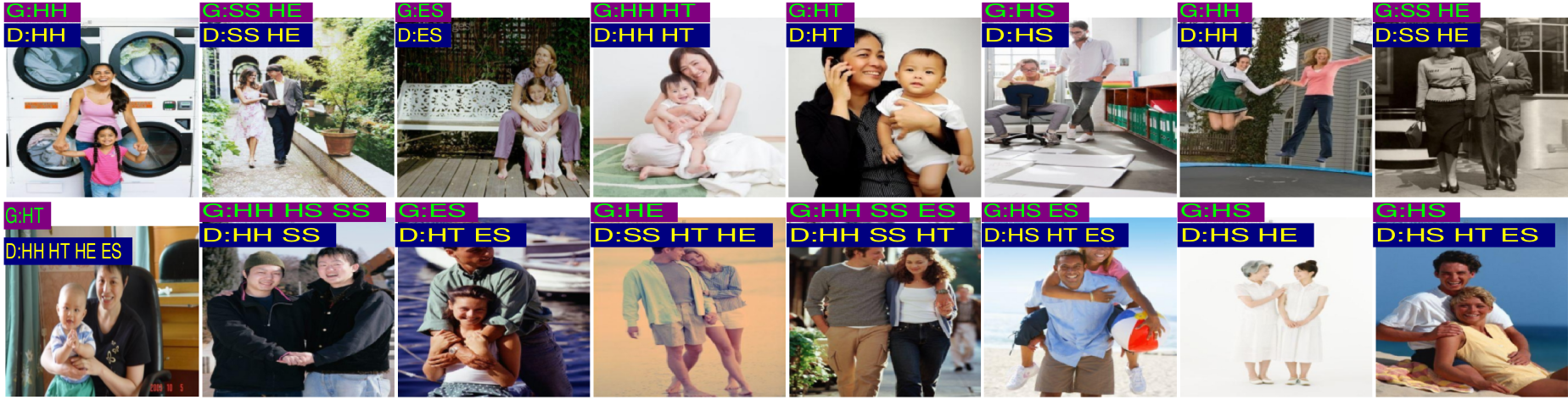}%
\linebreak
\begin{tabular}{ |c|c|c|c|c|c|c|c|c|c| }
\hline
  & HH & HS & SS    & HT    & HE & ES & Mean(a) & Mean(b)\\
\hline
Ours                 & \textbf{59.7} &  \textbf{52.0} & 53.9  &  33.2  &  36.1  &  36.2 &  \textbf{45.2} & \textbf{47.58}  \\
\hline
\cite{uci-dataset}   & 37            &  29            &  50           &  \textbf{61} & 38 & 34 & 42 & 38  \\
\hline
\cite{visual-phrase} & 31           &20               & 40            &  20 & 11 & 12 & 22 & 23 \\
\hline
\cite{chu}           & 41.2         &35.4             & \textbf{62.2}  & NA  & \textbf{43.9} & \textbf{55.0}  & NA & 47.54 \\
\hline
\end{tabular}
\caption{ 
\small
 Sample proxemics recognition. Row one: Our results (D) match the ground truth (G). Row two: Failure cases.
The table shows the
average precision in proxemics recognition. Mean(a) is the average of all classes. Mean(b) excludes class HT.
Numbers are percentages.
}
\label{fig:prox}
\end{figure}

\noindent \textbf{Application for proxemics recognition:}
Finally, we demonstrate the utility of our human region parsing for \emph{proxemics} recognition.  
Proxemics is the study of the spatial separation individuals naturally maintain in social situations.  
The UCI dataset was created to study proxemics, and is labeled for 6 classes:
 hand-hand (HH),
hand-shoulder (HS), shoulder-shoulder (SS), hand-torso (HT), hand-elbow (HE)
and elbow-shoulder (ES).

We use features that include the min and max
distances between each pair of upper body part regions
of a person pair normalized by the average scale of the two subjects, the normalized horizontal
and vertical distance of heads and the scale difference.
The data for training and testing
are uniformly split at random, following the setup in~\cite{uci-dataset}.
To learn the 6 proxemics classes on top
of these features,
we use a random forest classifier with 100 trees and unlimited tree depth.
We repeat the experiment 10 times and report the average accuracy.
We do not use ground truth head locations.

Fig.~\ref{fig:prox} shows sample classifications and AP scores.
Our average AP score is higher than all the competing methods~\cite{uci-dataset,visual-phrase,chu}.  
Our weakness vs.~\cite{uci-dataset} on HT is likely because not only baby hugging but also
other hand-on-torso images are classified as HT.
Compared to the prior pose detectors, our method is more resistant to large occlusion, 
non-pedestrian poses, and complex interaction among
   people.   We identify the membership of
   each pixel to each human instance; local errors, e.g. broken regions, do not cause a big problem.

\section{Conclusion}
We propose a novel method to segment human instances and label their body parts using region assembly.
The proposed method is able to handle complex human interactions,
occlusion, difficult poses, and is rotation and scale invariant.
Our branch and bound method is fast and gives reliable results.
Our method's results compare favorably to a wide array of alternative methods.

\section{Acknowledgments}
This research is supported in part by U.S. NSF 1018641 and a gift from Nvidia (HJ)
and ONR PECASE N00014-15-1-2291 (KG).

\appendix
\section{Appendix}
\subsection{Details on human body part CNN,
head candidate detection and obtaining 
semantic 
segmentation map using graph cuts
} \label{app:cnn}
We train the semantic segmentation body part CNN on the
LSP dataset~\cite{lsp} with 27 classes (including body joints,
in-between points and background class) using Caffe \cite{caffe}.
During training, the CNN's input size is 55$\times$55
and the first convolution layer has stride 1 instead of 4. The fully
connected layers before the output are replaced by simpler LeNet fully connected
layers and then converted to fully convolutional networks to
facilitate pixel-level semantic segmentation.
The training image
patches are rotated and flipped so that the trained classifier is
rotation invariant. When testing, all the images are rescaled so that
the longest edge is 500-pixels; the CNN input size is the same as the input image.
The 27 classes are finally merged to 4 part types---head, torso, arm and leg---
and a background class
when obtaining the soft semantic segmentation.

For each input image, we compute the CNN soft semantic maps of the five classes
(arm, leg, torso, leg and background) in different scales.
For each scale level, the input image is downsized to a specific resolution.
The downsizing factors range
from $100\%$ to $25\%$ with the step of $5\%$.
There are totally 16 different scale levels.
The soft semantic map of each class is then generated using
max-pooling. The soft semantic maps of the five classes are
finally normalized so that each value represents a probability.

The head points are localized on the merged soft head map by non-max suppression.
The non-maximum suppression uses the window of $7\times7$ and the
threshold of 0.2.
After each head point is localized, we find the head's scale by
going through the stack of head maps with
different scales
and finding the map that gives the largest value at the head point.
We record the head scale as the inverse of the corresponding image downsizing factor.
For instance, if at one head point, the semantic map with the downsizing factor of 0.5
gives the largest response in the CNN head maps, the head candidate has
a scale 2.

We compute the final semantic segmentation map using the graph cuts
method, which minimizes
the assignment of five labels: head, torso, arm, leg and background
to each pixel in the input image.
Two passes of graph cuts are used.
In the first pass, we label each image pixel as human foreground (body part pixel) or
background (non-human pixel). The unary term is one minus the pixel level
probability of these two classes from the CNN.
The weights are 1 and 0.2 for the unary term and binary term respectively.
The binary term smooths the labeling between neighboring pixels.
Each neighbor pixel pair contributes to the binary term; the contribution
equals the weight times the distance between the two neighbor pixels' labels.
We define the distance between two labels to be zero if they are the same and otherwise
the distance is one.
In the second pass, we only label the foreground pixels determined
from the first pass as the four body part classes: head, arm, leg or torso.
We also use one minus the pixel probability from the CNN in the unary term and the same
setting
in the pairwise term as pass one.
The weights are also 1 and 0.2 for the unary and binary terms.
To improve the result,
we include a term that penalizes pixels to be labeled
as arm or torso if they fall out of the arm range radius of each detected head
in the image. The arm range radius is computed using the head's scale and
 a 150-pixel tall reference person's maximum head to hand distance.
The penalty is applied in the alpha-expansion procedure.
For a pixel out of the range radiuses of all the detected heads,
 a penalty of 1 is included
in the unary term of the alpha node if the alpha label is torso or arm, or
a penalty of 1 is included in the unary term of the pixel node
if the current pixel label is torso or arm.
The first pass graph cuts can be optimized using the maxflow method.
In the second pass, 20 iterations are used in the alpha-expansion.

We use a push-relabel maxflow algorithm in the implementation.
The maxflow algorithm requires integer edge capacities;
we convert the floating point capacities in the above formulation to integers
by multiplying 1000 and then taking the floor.

\subsection{Parameter optimization} \label{app:para}
With enough labeled data, we can automatically set the parameters in
the the proposed method by
maximizing the margin on positive/negative examples via a linear program.
$\varepsilon$ can be obtained from the color statistics of the ground truth data.
We assume $\tau$ fixed for now.
We represent all the other weight parameters as a vector $v$.
Let
$w^i_p$ be the values of different terms weighted by these coefficients for the positive segmentation
sample of
image $i$, and $w^{i,j}_n$ be the corresponding values of the $j$th negative segmentation sample of image $i$.
Positive segmentation exemplars are from ground truth labeling.
Negative exemplars can be randomly generated using region proposals, which satisfy the overlap, color exclusion
and unique assignment constraints.
We maximize the energy difference between the positive and negative training examples using a similar formulation
to \cite{hoiem}:
\begin{align*}
&\max \sum_i u_i \;\;\; \mbox{s.t.} \; v^T(w^{i,j}_n-w^{i}_p) > u_i, \forall i,j, \; v^T1=1, \; v \ge 0.
\end{align*}
The optimization can be solved efficiently using linear programming.
$\tau$ has a small range and can be found by exhaustive search to achieve the highest segmentation performance
on the training dataset.

\end{document}